%% file: main.tex
\documentclass[sigconf]{acmart}

\AtBeginDocument{%
  }

\copyrightyear{2026}
\acmYear{2026}
\setcopyright{cc}
\setcctype{by-nc-nd}
\acmConference[KDD '26]{Proceedings of the 32nd ACM SIGKDD Conference on Knowledge Discovery and Data Mining V.2}{August 09--13, 2026}{Jeju Island, Republic of Korea}
\acmBooktitle{Proceedings of the 32nd ACM SIGKDD Conference on Knowledge Discovery and Data Mining V.2 (KDD '26), August 09--13, 2026, Jeju Island, Republic of Korea}
\acmDOI{10.1145/3770855.3819015}
\acmISBN{979-8-4007-2259-2/2026/08}

\usepackage{booktabs}
\usepackage{multirow}
\usepackage{array}
\usepackage{graphicx}
\usepackage{subcaption}
\usepackage{balance}

\begin{document}

\title{Beyond Time Series: Spatial Reasoning for Epidemic Forecasting via Multimodal Learning}

\author{Diana Guadalupe Gomez}
\email{dggomez@umich.edu}
\affiliation{%
  \institution{University of Michigan}
  \city {Ann Arbor}
  \state{MI}
  \country{USA}
}

\author{Chenwei Wu}
\email{chenweiw@umich.edu}
\affiliation{%
  \institution{University of Michigan}
  \city {Ann Arbor}
  \state{MI}
  \country{USA}
}

\author{Zhiyi Wang}
\email{jerrywzy@umich.edu}
\affiliation{%
  \institution{University of Michigan}
  \city {Ann Arbor}
  \state{MI}
  \country{USA}
}

\author{Liyue Shen}
\email{liyues@umich.edu}
\affiliation{%
  \institution{University of Michigan}
  \city {Ann Arbor}
  \state{MI}
  \country{USA}
}

\author{Alexander Rodríguez}
\email{alrodi@umich.edu}
\affiliation{%
  \institution{University of Michigan}
  \city {Ann Arbor}
  \state{MI}
  \country{USA}
}

\renewcommand{\shortauthors}{Diana Guadalupe Gomez, Chenwei Wu, Zhiyi Wang, Liyue Shen, and Alexander Rodríguez}

\begin{abstract} 
Epidemic forecasting models typically rely on surveillance data reported over administrative regions, treating them as atomic units, thereby obscuring sub-regional spatial structure that shapes disease dynamics. We introduce a spatially structured multimodal epidemic forecasting setting that integrates region-level temporal surveillance data with spatially localized auxiliary signals that are misaligned in resolution and structure, reflecting realistic public health reporting constraints. Building on this formulation, we propose \textbf{M-SPICE} (\textbf{M}ultimodal \textbf{SP}at\textbf{I}al \textbf{C}ontext for \textbf{E}pidemic Forecasting), a structure-aware spatiotemporal forecasting framework that performs joint reasoning over temporal disease dynamics and spatial context via attention-based multimodal fusion, allowing spatial signals to selectively condition temporal representations across forecast horizons. We evaluate our approach on real-world COVID-19, influenza, and influenza-like illness (ILI) forecasting tasks under realistic real-time evaluation protocols. Across all forecasting settings, our method consistently outperforms state-of-the-art multivariate time-series, multimodal, and epidemiological forecasting baselines while maintaining strong probabilistic forecasting performance. Finally, interpretability analyses reveal when, where, and how spatial signals are leveraged, highlighting settings in which purely temporal, region-aggregated models are most likely to fail.
\end{abstract}




\maketitle

\input{1_intro}
\input{2_related}

\input{3_problem}

\input{4_method}
\input{5_data}
\input{6_results}

\section{Conclusion}
In this work, we study a realistic multimodal epidemic forecasting setting in which regional surveillance time series must be integrated with spatially localized auxiliary data that are misaligned in resolution and structure. We show that relying on temporal signals alone is insufficient for accurate medium-term forecasting, and that incorporating higher-resolution spatial context through structure-aware multimodal fusion leads to consistent performance gains across diseases and forecast horizons. We further demonstrate that these gains remain robust across diverse forecasting settings, with improvements becoming more pronounced at longer forecast horizons. Beyond improving forecast accuracy, our framework supports reliable probabilistic forecasting across COVID-19, influenza, and ILI targets, achieving a favorable balance between calibration and interval sharpness. Through interpretability and robustness analyses, we provide insight into when, where, and how spatial context contributes to epidemic forecasting. Statistical analyses reveal that spatial context is most beneficial in regions with greater cross-boundary interaction and fine-grained spatial heterogeneity, while attention-based analyses suggest that spatial information primarily serves to condition evolving temporal representations. Importantly, results from the Michigan case study demonstrate that the proxy spatial disease burden maps capture much of the value of disease-matched, temporally aligned spatial inputs, providing a practical alternative when such surveillance data are unavailable. Together, these results highlight the importance of modeling spatial structure explicitly and position our framework as an effective and flexible approach for multimodal epidemic forecasting under public health reporting constraints. 

\textit{Future work} will explore extending the framework to support multi-scale spatial reasoning under realistic public health reporting constraints. Emerging efforts such as the MetroCast Hub~\cite{metrocast} enable access to sub-state epidemiological surveillance data aggregated at the level of health service areas (HSAs), which are single- or multi-county regions reflecting healthcare-seeking patterns. Incorporating such signals would allow the model to reason across multiple geographic scales beyond state-level aggregation.

\section{Limitations and Ethical Considerations}

\paragraph{Data availability and realism.}
One limitation of our study is that, although we simulated a real-time forecasting environment in our experimental setup, we did not rely on vintage data—that is, the versions of the data that were available at each specific point in time. A key impediment was the lack of a repository of vintage county-level data; for example, CMU’s Delphi Data Repository provides vintage National Center for Health Statistics (NCHS) mortality data only at the state level. To avoid inconsistencies between state- and county-level datasets, we therefore opted to use revised data.
We believe future work can further address additional data-quality challenges inherent to real-time forecasting, such as backfill~\cite{kamarthi2021back2future}, anomalies that were only identified and curated retrospectively~\cite{altieri2021curating,rodriguez_deepcovid_2021}, and potential reporting biases~\cite{mhasawade2021machine}.\\
\indent\textit{Ethical considerations and data use.}
The primary experiments use aggregated, de-identified, publicly available surveillance data obtained from official public health sources. The Michigan case study additionally uses private de-identified county-level hospitalization data provided by the Michigan Department of Health and Human Services (MDHHS) under a data use agreement. The work does not involve human subjects or personally identifiable information and therefore does not raise concerns related to consent or individual privacy. As with all epidemic forecasting systems, model outputs should be interpreted as decision-support tools rather than definitive predictions in high-stakes public health settings.\\
\indent\textit{Fairness and representativeness.}
While our framework is designed to operate under realistic public health reporting constraints, its performance may vary across regions due to differences in surveillance quality, reporting practices, and population structure. We do not explicitly model or correct for such disparities in this work. Future extensions could investigate how spatial data quality and reporting heterogeneity influence forecast reliability and explore methods to improve robustness across diverse regions.


\begin{acks}
This publication was made possible by the Insight Net cooperative agreements NU38FT000002 from the CDC’s Center for Forecasting and Outbreak Analytics (CDC-RFA-FT-23-0069). Its contents are solely the responsibility of the authors and do not necessarily represent the official views of the Centers for Disease Control and Prevention. LS and CW acknowledge funding support by the National Science Foundation (NSF) via grant IIS-2435746, Defense Advanced Research Projects Agency (DARPA) under contract No. HR00112520042, as well as the University of Michigan MIDAS PODS Grant Award.
\end{acks}

\section{GenAI Disclosure}
Generative AI tools were used to assist with grammar correction and improve clarity of the paper. These tools were also used to help support tables and figure presentation and refinement.

\bibliographystyle{ACM-Reference-Format}
\bibliography{refs}

\appendix
\section{Data Preprocessing}
\textit{{Spatial data preprocessing.}}
Weekly county-level COVID-19 hospitalization maps were constructed from CDC-reported admissions per $100{,}000$ population over the preceding week. To reduce the effect of heavy-tailed hospitalization rates and extreme values, we applied a $\log(1+x)$ transformation, followed by global z-score standardization using training-set statistics.

\textit{{Temporal data preprocessing}}
The epidemiological time series were applied a z-score normalization based on training data statistics and constructed into fixed-length sliding windows, $L$ of weekly observations. This design preserves raw temporal structure while ensuring numerical stability across inputs.

\section{Training and Implementation Details}
\label{app:training_details}

All experiments are conducted on a single NVIDIA A100 GPU. The multimodal framework uses 8 data loading workers, while temporal-only models use 4 workers. 

\textit{{Auxiliary Spatial Pretraining.}}
The spatial encoder is pretrained once using an auxiliary reconstruction objective on county-level COVID-19 hospitalization maps. This pretraining step requires approximately 40 minutes and is performed only once. After pretraining, the encoder parameters are frozen and reused across all forecasting experiments and random seeds.

\textit{Forecasting Model Training.}
The forecasting model is trained independently for each random seed using the frozen spatial encoder. Training requires approximately 4 hours per seed. Models are optimized using Adam with early stopping.

\textit{Reproducibility.}
All experiments are conducted with fixed random seeds. The same training, evaluation, and preprocessing pipelines are used across all model variants and ablation settings.

\section{Ablation and Robustness Analyses}
\label{app:robustness}
\subsection{Spatial Input Ablation}
\label{app:spatial_ablation}
To further examine the role of sub-regional spatial context and its ability to generalize across diseases, we perform inference-time ablations on the spatial inputusing the final influenza forecasting models from Table~\ref{tab:results_grouped_by_disease}. We compare three settings:  (1) dynamic COVID-19 county-level hospitalization maps, (2) static COVID-19 maps fixed to a single representative week, and (3) no spatial input in Table~\ref{tab:flu_ablation}. All models are evaluated over the 2023W41--2024W12 period with model weights held fixed.

\begin{table}[htbp]
\centering
\captionsetup{skip=4pt}
\caption{Ablation of spatial inputs for influenza forecasting using a frozen model. Results are reported as mean NRMSE across regions for 1--4 week ahead predictions, averaged over seeds $= [5,17,33]$, where \emph{Avg} is the mean across horizons.}
\small
\label{tab:flu_ablation}
\setlength{\tabcolsep}{4pt}
\renewcommand{\arraystretch}{1.1}
\resizebox{0.9\linewidth}{!}{
\begin{tabular}{lccccc}
\toprule
\textbf{Input Setting} & 
\textbf{W1} & 
\textbf{W2} & 
\textbf{W3} & 
\textbf{W4} & 
\textbf{Avg} \\
\midrule

Dynamic COVID maps (Ours) &
\textbf{0.065} &
\textbf{0.099} &
\textbf{0.125} &
\textbf{0.140} &
\textbf{0.107} \\

Static COVID maps &
0.080 &
0.130 &
0.160 &
0.190 &
0.140 \\

No COVID maps &
0.181 &
0.133 &
0.162 &
0.182 &
0.164 \\

\bottomrule
\end{tabular}
}
\end{table}

Dynamic spatial inputs yield the best performance across all horizons, followed by static maps, while removing spatial inputs leads to the largest degradation. This ordering indicates that both the presence of spatial information and its time-varying structure contribute meaningfully to forecasting performance. The substantial performance drop when removing COVID-19 maps further suggests that the model leverages spatial representations learned from one disease to improve forecasting in another, supporting cross-disease generalization of spatial context.

\subsection{Regional Variation Analysis}
\label{app:pearson}
Table~\ref{tab:region_correlation} provides the correlation analysis relating regional structural characteristics to performance gains from spatial integration. 

\begin{table}[htbp]
\centering
\small
\captionsetup{skip=4pt}
\caption{Correlation between regional structural characteristics and performance gains from spatial integration. Pearson correlation coefficients ($r$) and $p$-values are reported for average performance across forecast horizons.}
\label{tab:region_correlation}
\resizebox{0.7\linewidth}{!}{
\begin{tabular}{lcccc}
\toprule
\textbf{Predictor} 
& \multicolumn{2}{c}{\textbf{COVID}} 
& \multicolumn{2}{c}{\textbf{ILI}} \\
\cmidrule(lr){2-3} \cmidrule(lr){4-5}
& $r$ & $p$ & $r$ & $p$ \\
\midrule
County count & +0.300* & 0.034 & +0.430** & 0.002 \\
Border count & +0.551** & $<0.001$ & +0.352* & 0.012 \\
\bottomrule
\end{tabular}
}
\end{table}
\vspace{-5pt}

\section{Ablation on Model Components}
\label{app:model_ablation}
Ablations are conducted on spatial representations, fusion depth, and gating design, as these components are central to modeling structured cross-modal interactions under resolution misalignment.

\subsubsection{Spatial Encoder Pretraining}
The effect of spatial encoder pretraining is evaluated by varying both the input modalities and reconstruction objectives, and measuring downstream forecasting performance on COVID-19 and Flu hospitalization tasks.

\textit{{Effect of input modality.}}
We first compare encoders pretrained using disease-burden maps alone (COVID) versus jointly with temperature (COVID + Temp), while keeping the reconstruction objective fixed to COVID. Table~\ref{tab:ablation_pretraining} shows that incorporating temperature during pretraining yields very similar performance to using COVID alone for COVID forecasting. Similarly, in the flu transfer setting (Table~\ref{tab:pretraining_ablation_flu}), including temperature produces small and inconsistent changes across horizons, with no improvement in overall performance. These results suggest that temperature provides limited additional information beyond that captured by proxy maps.

\begin{table}[htbp]
\centering
\small
\setlength{\tabcolsep}{4pt}
\renewcommand{\arraystretch}{1.1}
\captionsetup{skip=4pt}
\caption{Ablation on spatial encoder pretraining for COVID-19 hospitalization forecasting. We compare input modalities and reconstruction objectives. Results are reported as mean NRMSE over seeds $= [5,17,33]$.}
\label{tab:ablation_pretraining}
\resizebox{0.9\linewidth}{!}{
\begin{tabular}{ll|ccccc}
\toprule
\textbf{Input} & \textbf{Reconstruction} 
& \textbf{W1} & \textbf{W2} & \textbf{W3} & \textbf{W4} & \textbf{Avg} \\
\midrule

COVID & COVID 
& 0.167 & 0.167 & \textbf{0.207} & \textbf{0.235} & 0.194 \\

COVID + Temp & COVID 
& \textbf{0.164} & \textbf{0.167} & 0.208 & 0.236 & \textbf{0.194} \\
\midrule

\multicolumn{2}{l|}{Random Init (no pretraining)} 
& 0.181 & 0.180 & 0.233 & 0.273 & 0.217 \\
\bottomrule
\end{tabular}
}
\end{table}

\textit{{Effect of reconstruction objective.}}
We compare COVID-only reconstruction with joint reconstruction of COVID and temperature to assess whether preserving environmental information improves downstream performance. Table~\ref{tab:pretraining_ablation_flu} shows that joint reconstruction introduces modest horizon-dependent effects, slightly degrading performance at shorter horizons (W1--W2) while improving or maintaining performance at longer horizons (W3--W4). Although these changes result in comparable overall performance, the gains at longer horizons suggest that temperature may provide complementary signals for medium-term forecasting. More temporally resolved environmental data may further strengthen this effect.

\begin{table}[htbp]
\centering
\small
\captionsetup{skip=4pt}
\caption{Effect of input modality and reconstruction objective on downstream influenza forecasting. Results are reported as mean NRMSE over seeds $= [5,17,33]$.}
\resizebox{\linewidth}{!}{
\begin{tabular}{llccccc}
\toprule
\textbf{Input} & \textbf{Reconstruction} & \textbf{W1} & \textbf{W2} & \textbf{W3} & \textbf{W4} & \textbf{Avg} \\
\midrule
COVID & COVID 
& \textbf{0.144} & 0.199 & 0.230 & \textbf{0.283} & \textbf{0.219} \\
COVID + Temp & COVID 
& 0.149 & \textbf{0.196} & 0.249 & 0.287 & 0.220 \\
COVID + Temp & COVID + Temp 
& 0.151 & 0.201 & \textbf{0.247} & 0.283 & 0.220 \\
\bottomrule
\end{tabular}
}
\label{tab:pretraining_ablation_flu}
\end{table}
\balance
\textit{{Impact of pretraining.}}
Across all configurations, pretrained encoders substantially outperform random initialization (Table~\ref{tab:ablation_pretraining}), confirming that spatial pretraining is critical for learning useful representations. Notably, the gains from pretraining are driven by disease-burden structure, rather than auxiliary climate signal.

\subsubsection{Fusion Depth}
Inspired by public health analysis, we use a shallow fusion design in which spatial context refines temporal representations. To evaluate this choice, we compare models with one to three layers of cross-modal fusion on the COVID task.

\begin{table}[htbp]

\centering
\small
\setlength{\tabcolsep}{3pt}
\captionsetup{skip=4pt}
\caption{Ablation on fusion depth for cross-modal interaction. Results are reported as mean NRMSE over seeds $= [5,17,33]$ for COVID-19 hospitalization forecasting.}
\label{tab:ablation_fusion_depth}
\resizebox{0.6\linewidth}{!}{
\begin{tabular}{l|ccccc}
\toprule
\textbf{Fusion Depth} & \textbf{W1} & \textbf{W2} & \textbf{W3} & \textbf{W4} & \textbf{Avg} \\
\midrule
1-layer (Ours) & \textbf{0.164} & \textbf{0.167} & \textbf{0.209} & \textbf{0.236} & \textbf{0.194} \\
2-layer        & 0.180 & 0.176 & 0.212 & 0.239 & 0.202 \\
3-layer        & 0.173 & 0.175 & 0.215 & 0.245 & 0.202 \\
\bottomrule
\end{tabular}
}
\end{table}
As shown in Table~\ref{tab:ablation_fusion_depth}, a single fusion layer achieves the best performance across all forecast horizons. The performance gap is largest at shorter horizons (W1), where temporal dynamics dominate, suggesting that deeper fusion introduces unnecessary complexity without improving predictive performance. These results support the use of shallow fusion for incorporating spatial context.

\begin{table}[htbp]
\centering
\small
\captionsetup{skip=4pt}
\caption{Ablation on gating mechanisms for COVID-19 hospitalization forecasting. Results are reported as mean NRMSE over seeds $= [5,17,33]$.}
\label{tab:ablation_gating}
\resizebox{0.8\linewidth}{!}{
\begin{tabular}{l|ccccc}
\toprule
\textbf{Gating Mechanism} & \textbf{W1} & \textbf{W2} & \textbf{W3} & \textbf{W4} & \textbf{Avg} \\
\midrule
Horizon-dependent (Ours) & \textbf{0.164} & \textbf{0.167} & \textbf{0.209} & \textbf{0.236} & \textbf{0.194} \\
Single learned weight    & 0.169 & 0.172 & 0.211 & 0.240 & 0.199 \\
Uniform combination      & 0.174 & 0.175 & 0.218 & 0.249 & 0.204 \\
\bottomrule
\end{tabular}
}
\end{table}

\subsubsection{Horizon-Dependent Gating}
We compare our horizon-dependent gating with simpler alternatives, including a single learned weight and a uniform combination. Table~\ref{tab:ablation_gating} shows that the horizon-dependent gating mechanism achieves the best performance across all forecast horizons. Performance degrades as the gating mechanism becomes less flexible, indicating that the optimal contribution of spatial context varies across forecast horizons. These findings support our hypothesis that temporal and spatial information play different roles over the forecast trajectory and that horizon-specific weighting enables more effective multimodal integration.
\end{document}

%% file: 1_intro.tex
\section{Introduction}
\label{sec:intro}
\let\thefootnote\relax\footnotetext{
Code available at \url{https://github.com/complex-ai-lab/m-spice}.
}
The ability to accurately forecast epidemic trajectories is a cornerstone of modern public health, yet it remains challenged by a fundamental gap between disease transmission processes and data reporting practices. Infectious diseases evolve through biological, social, and environmental processes that vary at fine spatial scales within administrative regions~\cite{holmdahl2020wrong,marathe2013computational}, while epidemiological surveillance data are typically reported at coarser, administratively defined levels such as states or regions~\cite{reinhart2021open,rodriguez2024machine}. As a result, the fine-scale processes underlying disease transmission must be collapsed into coarse aggregates to match the spatial resolution of the available surveillance data.

\begin{figure}[t]
\centering
\includegraphics[width=\linewidth]{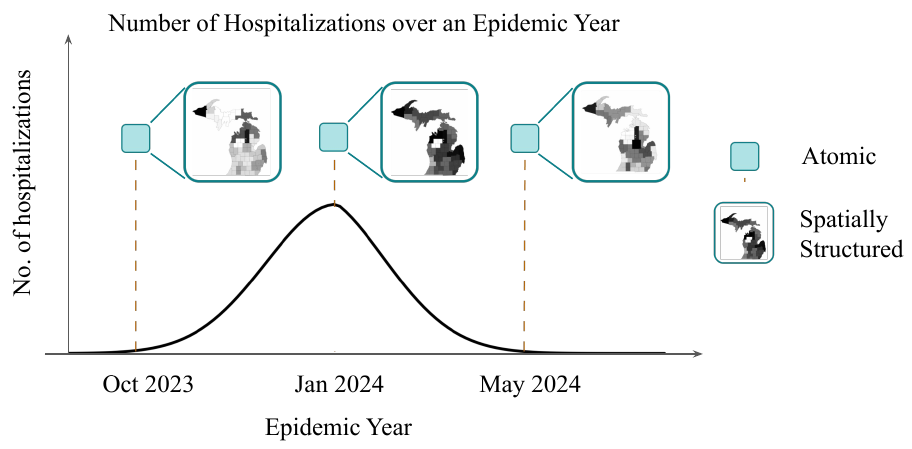}
\captionsetup{skip=8pt}
\caption{Atomic vs. spatially structured representations for epidemic forecasting. Atomic representation collapses each administrative region into a single temporal signal, whereas spatially structured preserve within-region heterogeneity.}
\label{fig:atomic_vs_structure}
\vspace{-15pt}
\end{figure}

These realities have given rise to a dominant modeling paradigm in which regions (such as states) are treated as spatially homogeneous, indivisible units of analysis. We refer to this paradigm as \emph{atomic} modeling (Figure~\ref{fig:atomic_vs_structure}).  Purely temporal multivariate time-series models~\cite{wang2023timemixer,zhang2023crossformer} operate on spatially aggregated regional signals, while epidemiology-informed models based on graph neural networks (GNNs)~\cite{deng2020cola,wu2018deep} represent regions as discrete nodes and model interactions between administratively defined units. As a result, such formulations are inherently boundary-locked: spatial relationships are encoded at the level of reporting boundaries, rather than reflecting sub-regional structure within regions. By collapsing each region into a single temporal signal, atomic models obscure sub-regional variation and limit the ability to investigate how localized spatial structure, such as environmental gradients or disease hotspots, shapes epidemic evolution over time.

In this work, we argue that overcoming the limitations of atomic modeling requires leveraging auxiliary modalities that provide structured spatial context to enrich coarse public health signals.
To this end, we introduce \textbf{M-SPICE} (\textbf{M}ultimodal \textbf{SP}at\textbf{I}al \textbf{C}ontext for \textbf{E}pidemic Forecasting), a structure-aware multimodal forecasting framework that integrates state-level temporal surveillance data with fine-resolution auxiliary spatial signals. Motivated by epidemiological evidence highlighting the role of temperature in modulating multiple aspects of disease dynamics~\cite{dalziel2018urbanization,soebiyanto2010modeling}, we begin by examining land surface temperature. We leverage dynamic temperature fields to capture sub-regional environmental structure; however, temperature alone is insufficient to characterize how such environmental variation translates into disease spread and severity. Therefore, we additionally incorporate proxy spatial distributions of hospitalizations at the county level, derived from historical records, which provide complementary context for interpreting the influence of environmental signals on regional epidemic trajectories. 

We can summarize our contributions as follows:
\begin{itemize}
    \item \textbf{Structure-aware multimodal epidemic forecasting.}
    We formulate a multimodal epidemic forecasting setting that moves beyond atomic regional representations by enabling sub-regional spatial context to inform regional predictions under realistic reporting constraints. To address this setting, we introduce \textbf{M-SPICE}, a structure-aware spatiotemporal framework that integrates temporal disease dynamics with spatial context via attention-based fusion.
    \item \textbf{Interpretability and epidemiological insight.}
    We provide a systematic analysis of when, where, and how spatial context improves epidemic forecasting across diseases, geographic regions, and forecast horizons. Through statistical analysis, attention-based interpretability methods, and ablation studies, we show that spatial context is most beneficial at longer forecast horizons and in regions with greater spatial heterogeneity. Our analyses further suggest that localized environmental and disease burden structure primarily serves to contextualize evolving temporal disease trajectories, demonstrating the framework’s potential utility as an exploratory tool for epidemiological analysis.
    \item \textbf{Multi-disease evaluation against strong baselines.}
    Across COVID-19, influenza, and ILI forecasting tasks on real-world public health data, our approach consistently surpasses multivariate time-series, multimodal, and epidemic forecasting baselines. All experiments follow realistic real-time forecasting settings, aligning with best practices for epidemic forecast evaluation and demonstrating the practical value of structure-aware multimodal fusion.
\end{itemize}

The remainder of the paper is organized as follows. We first review related work, then introduce our spatially structured multimodal problem formulation. We next describe the data and proposed methodology, present experimental results, and conclude with a discussion of implications and limitations.

%% file: 2_related.tex
\section{Related Work}
\label{sec:relatedworks}

\subsection{Epidemic Forecasting}

In disease spread modeling, approaches can be broadly categorized into three classes: mechanistic models, data-driven models, and hybrid methods that combine elements of both~\cite{rodriguez2024machine}.
\subsubsection{Mechanistic models}
Mechanistic models explicitly encode the causal mechanisms governing epidemic spread and have long served as a foundation for epidemiological analysis and forecasting. Classical compartmental models, such as SIR~\cite{hethcote_mathematics_2000} and SEIR~\cite{wu2020nowcasting}, describe disease dynamics through systems of differential equations that track population-level transitions between health states. Extensions of these models include metapopulation frameworks~\cite{balcan2009multiscale,pei2018forecasting}, which introduce additional heterogeneity by partitioning populations into interacting subpopulations, often defined by geographic units. At an even finer level of granularity, agent-based models explicitly simulate individual-level interactions~\cite{hinch2021openabm,chopra2023differentiable}.

A substantial body of work has sought to extend mechanistic models by incorporating multiple spatial scales (e.g., county-, state-, and national-level data)~\cite{gopalakrishnan2020globally,chinazzi2024multiscale,osthus2021multiscale} as well as exogenous drivers such as weather and environmental factors~\cite{shaman2009absolute,shaman2010absolute}. While these extensions can yield valuable epidemiological insight and are often highly interpretable, mechanistic models face well-documented challenges. In practice, they require extensive calibration, rely on strong structural assumptions, and can be overly rigid when confronted with noisy, sparse, or rapidly shifting real-world data~\cite{nsoesie2014systematic,holmdahl2020wrong,ye2025integrating}. Moreover, integrating non-traditional, high-dimensional, or multimodal data sources, such as remotely sensed environmental data, remains difficult within purely mechanistic formulations.

\subsubsection{Data-driven and Hybrid Models}
Early data-driven deep learning methods focused on identifying recurrent patterns in epidemic time series by comparing the similarity of temporal segments across time~\cite{adhikari2019epideep,kamarthi2021epifnp} or across locations~\cite{jin2021inter}. Building on these ideas, most spatiotemporal models adopted graph-based formulations, representing regions as nodes and encoding interactions through predefined or learned adjacency structures~\cite{deng2020cola,wu2018deep,Wan_EpiODE_ICML25}. Hybrid models seek to bridge the gap between the domain knowledge embedded in mechanistic formulations and the predictive flexibility of machine learning. A common strategy is to use deep learning backbones augmented with epidemiological priors, incorporating components of mechanistic models through knowledge-guided or physics-informed architectures and loss functions~\cite{rodriguez2023einns,wang2022causalgnn}. These approaches aim to improve generalization and interpretability while retaining the capacity to fit complex real-world data.

Despite these advances, most existing data-driven and hybrid models operate at coarse spatial resolutions and treat regions as homogeneous atomic units, typically at the national- or state-level, as discussed in Section~\ref{sec:intro}. While effective for capturing population-level transmission dynamics, such formulations rely on predefined state representations or region-level aggregation, which limits their ability to exploit sub-regional spatial heterogeneity and integrate spatially localized signals observed at higher resolutions.

\subsection{Deep Spatiotemporal Forecasting}
\subsubsection{Grid-based spatiotemporal models.} A large body of work in spatiotemporal learning has focused on modeling complex dynamics over space and time using either grid-based or graph-based representations. Grid-based approaches, including convolutional, recurrent, and attention-based architectures, model spatiotemporal processes on regular lattices and assume consistent spatial resolution across time and space. Early work such as ConvLSTM~\cite{shi2015convlstm} extended convolutional recurrent models to capture spatiotemporal dependencies on grids, while recent work such as Earthformer~\cite{gao2022earthformer} leverages spatiotemporal attention over grid tokens to capture long-range dependencies and produce dense spatial outputs at inference.

\subsubsection{Graph-based spatiotemporal models.}
Complementary to grid-based formulations, a substantial body of work models spatiotemporal dynamics using graph-based representations, where nodes correspond to spatial units and edges encode spatial proximity, mobility patterns, or learned inter-node dependencies~\cite{yu2018spatio, wu2020connecting, gao2021stan}.  
These approaches enable joint modeling of temporal dynamics and relational structure, and have demonstrated strong performance across a range of forecasting tasks. However, neither grid-based nor graph-based formulations are naturally suited to integrate spatially localized signals that are misaligned in resolution and structure.

\subsection{Multimodal Time Series Modeling}

Recent work has explored \emph{multimodal} time series modeling by combining temporal signals with auxiliary modalities such as images, text, and structured EHR streams. Early approaches addressed irregularly sampled multimodal data for clinical prediction~\cite{zhang2023improving}, while more recent methods have investigated disentangling stable and evolving image representations and integrating them with temporal streams through local and global fusion mechanisms~\cite{liu2025multimodal}. Other frameworks employ symbolic tokenization and mixture-of-experts routing to handle missing modalities and varying information quality~\cite{maestro2025}. While these methods demonstrate the effectiveness of multimodal fusion for temporal prediction, they generally treat modalities as complementary signals within a shared prediction task. In contrast, epidemic forecasting requires integrating modalities that differ fundamentally in both spatial resolution and functional role: regional surveillance time series provide the primary forecasting signal, whereas spatial fields or sub-regional disease burden maps provide contextual information that conditions predictions. This cross-resolution structural mismatch is not addressed by existing multimodal time-series frameworks and motivates our structure-aware formulation for incorporating spatial context into epidemic forecasting.

%% file: 3_problem.tex
\section{Problem Formulation}
\label{sec:problem}
We study the task of epidemic forecasting over a fixed short-term prediction horizon using heterogeneous temporal and spatial data sources collected across regions and time. Unlike prior formulations that operate solely on aggregated regional time series, we explicitly consider forecasting under heterogeneous and spatially misaligned contextual signals, reflecting realistic public health data collection and reporting constraints.

\par\noindent\textbf{\underline{Given:}}
Let $r \in \mathcal{R}$ index regions and $t \in \mathcal{T}$ index epidemiological weeks. For each region $r$ at time $t$, we observe the multivariate epidemiological time series $\mathbf{x}$ and the corresponding epidemic outcome $\mathbf{y}$ up to time $t$,
\[
\mathbf{x}_{r,1:t} \in \mathbb{R}^{t \times d_x},
\qquad
\mathbf{y}_{r,1:t} \in \mathbb{R}^{t},
\]
where $d_x$ denotes the number of epidemiological features. In addition, we observe time-aligned spatial context maps
\[
\mathbf{I}_{r,1:t} \in \mathbb{R}^{t \times C \times h \times w},
\]
where $C$ denotes the number of channels while $h$ and $w$ denote the dimensions of the spatial context maps. These maps provide complementary spatial information, such as environmental conditions or localized disease burden, at resolutions that are not aligned with administrative reporting units.

Existing epidemic forecasting formulations typically operate on region-level aggregates and treat each region as a homogeneous atomic unit, reducing potentially heterogeneous within-region disease patterns into a single temporal signal (Figure~\ref{fig:atomic_vs_structure}). In contrast, our formulation represents each region using spatially structured contextual information, preserving within-region heterogeneity that may influence disease dynamics and forecasting performance.

\par\noindent\textbf{\underline{Predict:}}
The goal is to forecast future epidemic targets
\[
\mathbf{y}_{r,t+1:t+H}
\]
over a prediction horizon of $H=4$ weeks, consistent with CDC forecasting initiatives that emphasize short-term epidemic prediction~\cite{cramer2021evaluation,reich2019accuracy}. The central challenge lies in integrating evolving temporal disease dynamics with spatial context whose relevance varies across regions and epidemic phases, and whose spatial structure is not aligned with epidemiological reporting units.

%% file: 4_method.tex
\section{Methodology}
\label{sec:method}

\begin{figure*}[t!]
\centering
\includegraphics[width=\linewidth]{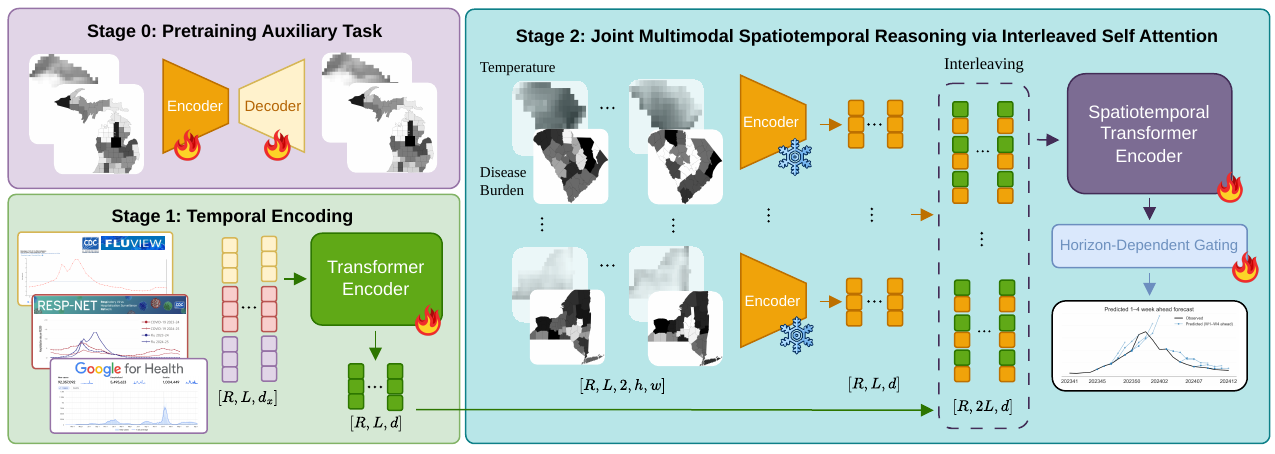}
\caption{Structure-aware multimodal epidemic forecasting pipeline. Stage 0 pretrains a spatial encoder via an auxiliary reconstruction task on temperature and disease burden maps. Stage 1 encodes regional epidemiological time series into temporal representations using a transformer-based encoder. Stage 2 performs joint multimodal spatiotemporal reasoning by interleaving temporal and spatial tokens prior to self-attention, followed by a horizon-dependent gating to generate multi-horizon epidemic forecasts.}
\label{fig:pipeline}
\vspace{-5pt}
\end{figure*}

We propose a structure-aware multimodal spatiotemporal forecasting model that integrates coarse temporal epidemiological dynamics with high-resolution auxiliary signals. The design is explicitly informed by how public health analysts reason about epidemic trajectories: recent temporal trends form the primary basis for short-term predictions, while spatial context—such as temperature gradients and sub-regional disease hotspots—is selectively incorporated to refine those trends over longer horizons. By interleaving temporal and spatial tokens prior to joint self-attention and horizon-dependent gating, the framework avoids the "atomic" limitation of treating states as indivisible nodes, instead allowing localized spatial features to dynamically condition regional trajectories. 

Our framework consists of three core components (Figure~\ref{fig:pipeline}): 
\begin{itemize}
    \item[(1)] a shared spatial encoder that maps temperature and county-level disease burden maps to time-aligned embeddings,
    \item[(2)] a temporal encoder that models recent epidemiological dynamics using a Transformer-based encoder,
    \item[(3)] a joint multimodal spatiotemporal reasoning module that enables cross-modal fusion via interleaved temporal and spatial tokens prior to joint self-attention, followed by a horizon-dependent gating mechanism for multi-step prediction.
\end{itemize}

\subsection{Shared Spatial Encoder for Temperature and Disease Burden Maps}
\label{sec:spatial_encoder}

Following the forecasting formulation, we use a fixed lookback window of length $L$. For each region $r$, the observed epidemiological history is represented as
\[
x_{r,t-L+1:t} \in \mathbb{R}^{L \times d_x}.
\]
To enrich the atomic regional signal, we incorporate sub-regional structure through auxiliary spatial modalities in the form of a 2-channel spatial tensor $\mathbf{I}_{r,t} \in \mathbb{R}^{2 \times h \times w}$ representing county-level disease burden maps and temperature fields. Because high-resolution disease-specific spatial surveillance is often unavailable, we use county-level COVID-19 hospitalization maps as a proxy measure of spatial disease burden across all forecasting targets. This formulation reflects realistic public health reporting constraints while providing geographically localized information that is not captured by aggregated regional surveillance signals.
 
For each time step in the lookback window, we encode spatial auxiliary information using the encoder component $f_{\mathrm{img}}$ of a pretrained autoencoder:

\[
\mathbf{u}_{r,t} = f_{\mathrm{img}}(\mathbf{I}_{r,t})
\in \mathbb{R}^{d_{\mathrm{img}}}.
\]

The encoder is frozen during forecasting training to preserve the spatial features learned during pretraining. We project these embeddings to form time-aligned spatial tokens:

\[
\mathbf{v}_{r,t} = \mathbf{W}_{\mathrm{img}}\mathbf{u}_{r,t} \in \mathbb{R}^{d}, \qquad
\mathbf{V}_{r} = [\mathbf{v}_{r,t-L+1}, \dots, \mathbf{v}_{r,t}] \in \mathbb{R}^{L \times d},
\]

\noindent where $W_{\mathrm{img}} \in \mathbb{R}^{d \times d_{\mathrm{img}}}$ is a learnable projection matrix. The projected sequence $\mathbf{V}_{r}$ provides a temporally aligned sequence of spatial embeddings that is used for downstream multimodal fusion. 
\\

\noindent\textbf{Auxiliary Task for Spatial Representation Learning.}
Stage 0 pretrains the spatial encoder using an auxiliary reconstruction task designed to capture epidemiologically meaningful spatial structure. During pretraining, the encoder maps the two-channel input
$\mathbf{I}_{r,t}
=
[\mathbf{I}^{\mathrm{temp}}_{r,t},
\mathbf{I}^{\mathrm{covid}}_{r,t}]
\in
\mathbb{R}^{2 \times h \times w}$
to a latent representation $\mathbf{u}_{r,t}$, which is passed to a decoder network $g_{\mathrm{dec}}$. The decoder reconstructs only the COVID hospitalization channel,
$\hat{\mathbf{I}}^{\mathrm{covid}}_{r,t}
=
g_{\mathrm{dec}}(\mathbf{u}_{r,t})$.
The auxiliary objective minimizes reconstruction error between the predicted and observed county-level COVID hospitalization maps. By conditioning reconstruction on both temperature and COVID inputs while supervising only the COVID channel, the auxiliary task is intended to encourage representations that incorporate environmental context when useful for reconstructing spatial disease burden patterns. Pretraining is performed prior to forecasting model training, after which encoder parameters are frozen and used to generate spatial embeddings for downstream forecasting.

\subsection{Temporal Encoder for Epidemiological Dynamics}
\label{sec:temporal_encoder}
Stage 1 encodes recent epidemiological dynamics using a Transformer temporal encoder. Given the multivariate epidemiological time series
$\mathbf{x}_{r,t-L+1:t} \in \mathbb{R}^{L \times d_x}$,
the sequence is processed by the encoder to produce temporally contextualized representations
\[
\mathbf{n}_r
=
f_{\mathrm{ts}}(\mathbf{x}_{r,t-L+1:t})
\in
\mathbb{R}^{L \times d_{\mathrm{ts}}}.
\]
These representations are then projected into the shared multimodal embedding space:
\[
\mathbf{M}_r = \mathbf{n}_r \mathbf{W}_{\mathrm{ts}} \in \mathbb{R}^{L \times d}, \qquad
\mathbf{M}_{r} = [\mathbf{m}_{r,t-L+1}, \dots, \mathbf{m}_{r,t}] \in \mathbb{R}^{L \times d},
\]
where $\mathbf{W}_{\mathrm{ts}} \in \mathbb{R}^{d_{\mathrm{ts}} \times d}$ is a learnable projection matrix. The projected sequence $\mathbf{M}_{r}$ is used for downstream multimodal reasoning.

\subsection{Joint Multimodal Spatiotemporal Reasoning via Interleaved Self-Attention}
\label{sec:fusion}
In Stage 2, temporal and spatial representations are integrated by interleaving temporal and spatial tokens chronologically to form a joint sequence:
\[
\mathbf{T}_r =
[\mathbf{m}_{r,t-L+1},
\mathbf{v}_{r,t-L+1},
\dots,
\mathbf{m}_{r,t},
\mathbf{v}_{r,t}]
\in
\mathbb{R}^{2L \times d}.
\]
Both temporal and spatial tokens are augmented with the same positional encoding corresponding to their time index prior to interleaving. Consequently, each temporally aligned token pair $(\mathbf{m}_{r,t-\ell}, \mathbf{v}_{r,t-\ell})$ shares an identical positional encoding, allowing the attention mechanism to interpret them as co-temporal views of the same epidemic state rather than unrelated sequence elements.

A Transformer encoder layer applies self-attention over $\mathbf{T}$ to produce fused representations $\tilde{\mathbf{T}}_r
= f_{\mathrm{joint}}(\mathbf{T}_r) \in \mathbb{R}^{2L \times d}$.
Interleaving preserves temporal alignment between modalities while allowing limited and symmetric information exchange. We intentionally restrict joint reasoning to a single attention layer to avoid excessive fusion and to mimic the selective contextual refinement in human-expert epidemiological analysis.

\subsection{Horizon-Dependent Gated Forecasting}
\label{sec:gating}
Final predictions are derived from temporally indexed representations after joint reasoning. From the joint output $\tilde{\mathbf{T}}_r$, we extract summary representations for forecasting. Let $\tilde{\mathbf{m}}_{t}$ denote the final temporal token and $\tilde{\mathbf{v}}_{t}$ denote the final spatial token. Predictions from the temporal and spatial representations are generated using separate prediction heads:
\[\hat{\mathbf{y}}^{\mathrm{ts}} = f_{\mathrm{ts}}^{\mathrm{out}}(\tilde{\mathbf{m}}_{t}), \qquad
\hat{\mathbf{y}}^{\mathrm{sp}} = f_{\mathrm{sp}}^{\mathrm{out}}(\tilde{\mathbf{v}}_{t}).\]
We combine these predictions using a learned horizon dependent gating mechanism
\[\hat{\mathbf{y}}_{t+1:t+H} = (1 - \boldsymbol{g}) \odot \hat{\mathbf{y}}^{\mathrm{ts}} + \boldsymbol{g} \odot \hat{\mathbf{y}}^{\mathrm{sp}},
\]
where $\boldsymbol{g} \in (0,1)^H$ is a learned vector controlling the contribution of spatial context at each forecast horizon.
This formulation enables short-term forecasts to rely primarily on temporal dynamics while allowing greater influence of spatial context at longer horizons.

\subsection{Training Objective}
\label{sec:training}
The forecasting model is trained to minimize the mean squared error between predicted outputs $\hat{\mathbf{y}}_{r,t+1:t+H}$ and ground-truth targets $\mathbf{y}_{r,t+1:t+H}$ over the forecast horizon $H$:
\[
\mathcal{L}_{\mathrm{forecast}}
=
\frac{1}{H}
\sum_{i=1}^{H}
\left(
\hat{y}_{r,t+i}
-
y_{r,t+i}
\right)^2 .
\]
The spatial encoder is pretrained using the auxiliary reconstruction objective described in Section~\ref{sec:spatial_encoder} and is frozen during forecasting model training. All other components are trained end-to-end.

%% file: 5_data.tex
\section{Multimodal Real-World Datasets}
\label{subsec:data}
We use multiple real-world data sources to capture complementary aspects of epidemic dynamics. For epidemiological outcomes, we use surveillance data from the Centers for Disease Control and Prevention (CDC), specifically the ILINet~\cite{cdc2025ilinet} and RESPNet~\cite{cdc2025respnet} systems. ILINet provides weekly reports of influenza-like illness (ILI) activity across the U.S. regions, reflecting state- and national-level transmission dynamics. RESPNet provides laboratory-confirmed influenza and COVID-19 hospitalization counts, which serve as a measure of disease severity. Thus, these data sources incorporate information about transmission intensity and severity over time. 

To incorporate early behavioral and syndromic signals, we additionally include symptom-related search trends from Google Symptom Search~\cite{bavadekar2020google}. From the approximately 400 symptoms tracked by the platform, we select those associated with ILI and COVID-like illness. These search trends provide a population-level information-seeking behavior and have been shown to offer early signals of changes in disease activity~\cite{reinhart2021open}. Google Symptom Search data are temporally aligned with epidemiological surveillance data and included as additional channels in the multivariate time-series inputs.

Spatial context is provided by near-surface air temperature derived from the ERA5 reanalysis dataset \cite{hersbach2023era5_cds}. In addition, we incorporate county-level COVID-19 hospitalization data obtained from the CDC. While temperature captures environmental conditions known to modulate respiratory virus transmission, COVID-19 hospitalization maps reflect contemporaneous respiratory disease burden. Together, these spatial signals provide complementary context by representing external environmental drivers and ongoing disease activity, respectively.

%% file: 6_results.tex
\section{Experiments}
\label{sec:experiments}
We evaluate the following research questions:
  \textbf{RQ1:} Does integrating spatial context improve forecasting performance and reliability compared to temporal-only models? \textbf{RQ2:} How do the benefits of spatial integration vary across diseases, regions, and forecast horizons? \textbf{RQ3:} What spatial context is leveraged by the model, and what does it reveal about patterns in disease dynamics?

\subsection{Protocol}
\textbf{Setup.}
We train and evaluate all models using a rolling-origin evaluation protocol consistent with CDC forecasting practice. Time is indexed using epidemiological weeks (e.g., 2020W43 denotes week 43 of 2020). Models are trained using historical observations from 2020W43 through 2023W39. Evaluation is performed over forecast origins spanning 2023W40--2024W12, where each model generates $H=4$ week-ahead predictions. Additional implementation details are provided in Appendix~\ref{app:training_details}.\\

\noindent\textbf{Metrics.} Forecast accuracy is measured using normalized root mean squared error (NRMSE) where values are normalized using min--max scaling per region
\[
\mathrm{NRMSE}
=
\sqrt{
\frac{1}{N}
\sum_{i=1}^{N}
\left(
\hat y_i^{\mathrm{norm}}
-
y_i^{\mathrm{norm}}
\right)^2
}.
\]
  Prediction uncertainty is assessed by applying Adaptive Conformal Inference (ACI)~\cite{NEURIPS2021_0d441de7} as a post-hoc calibration procedure and evaluating the resulting prediction intervals using empirical 90\% coverage (Cov90), which measures interval coverage calibration, and relative Weighted Interval Score (rWIS), which evaluates overall interval quality by balancing calibration and sharpness relative to a persistence baseline forecast, following~\cite{mathis2024flusight}. \\

\noindent\textbf{Baselines.}
We compare against a representative set of forecasting models spanning temporal-only methods, multimodal approaches, and epidemic forecasting models. While prior work has explored multimodal epidemic forecasting, existing approaches differ in the type of auxiliary information used and how it is integrated with epidemiological time series. In particular, they do not explicitly address the integration of spatially localized environmental and disease burden maps that are misaligned in resolution and structure with respect to their reporting units. Our baseline comparisons are therefore designed to contextualize the benefits of structure-aware spatial fusion under realistic public health reporting constraints.

\textit{Temporal-only forecasting baselines.}
We include classical and neural time-series models and provide epidemiological surveillance as inputs to assess the extent to which forecast accuracy can be achieved without spatial context. Specifically, we evaluate ARIMA~\cite{box2015time}, GRU~\cite{DBLP:journals/corr/ChoMBB14}, Autoformer~\cite{wu2021autoformer}, Crossformer~\cite{zhang2023crossformer}, Transformer-based architectures designed for long-horizon multivariate forecasting, and Timemixer~\cite{wang2023timemixer}, a recent state-of-the-art model that captures temporal patterns through multiscale mixing.

\textit{Multimodal forecasting baselines.}
We compare against two multimodal forecasting models. Specifically, we evaluate Maestro~\cite{maestro2025}, which was designed for multimodal fusion over heterogeneous temporal sensor sources, and DiPro~\cite{liu2025multimodal}, which was intended for integrating time series with medical imaging inputs (chest X-ray images). While these approaches leverage multimodal information, they were not designed to integrate time series with spatially localized environmental or disease burden maps under reporting-induced resolution and structural misalignment.

\textit{Epidemic forecasting baselines.} We additionally compare against forecasting models that are representative deep learning approaches for epidemic forecasting. These models use the same epidemiological surveillance data but do not ingest spatial maps. Specifically, we evaluate a persistence baseline used in the CDC epidemic prediction initiatives~\cite{cramer2021evaluation,mathis2024flusight}, which predicts future values using the most recently observed target; ColaGNN~\cite{deng2020cola}, which models spatial dependencies between regions using graph neural networks and recurrent temporal dynamics; and CNNRNNRes~\cite{wu2018deep}, which combines convolutional and recurrent neural networks to capture spatial and temporal disease dynamics. We also attempted to include EARTH~\cite{Wan_EpiODE_ICML25}, but were unable to due to unresolved implementation issues that prevented a fair and reproducible comparison.

\subsection{RQ1: Forecasting with Spatial Context}
\label{subsec:main_results}
We present the main quantitative results comparing performance across temporal-only, epidemic, and multimodal forecasting models using regional epidemiological surveillance data in Table~\ref{tab:results_grouped_by_disease}. Additional ablation studies analyzing spatial encoder pretraining, fusion design, and gating mechanisms are provided in Appendix~\ref{app:model_ablation}. All methods rely on the same surveillance inputs, while multimodal models additionally incorporate temperature and proxy maps. Our method demonstrates consistent gains from integrating spatial context through structure-aware multimodal fusion under realistic reporting and resolution constraints.

\begin{table*}[t]
\centering
\caption{Weekly forecasting results for 1--4 week ahead surveillance prediction across COVID-19 and influenza hospitalizations and ILI percentage.
Results are reported as mean NRMSE for seeds $= [5,17,33]$, where \emph{Avg} is the mean across horizons.
Models are trained on data from 2020W43--2023W40 and evaluated on 2023W41--2024W16.
Best performance is shown in bold.}
\label{tab:results_grouped_by_disease}
\vspace{-10pt}
\setlength{\tabcolsep}{5.75 pt}
\renewcommand{\arraystretch}{.58}
\small
\resizebox{\textwidth}{!}{%
\begin{tabular}{l|ccccc|ccccc|ccccc}
\toprule
\multirow{2}{*}{\textbf{Model}} &
\multicolumn{5}{c|}{\textbf{COVID hospitalizations}} &
\multicolumn{5}{c|}{\textbf{influenza hospitalizations}} &
\multicolumn{5}{c}{\textbf{ILI percentage}} \\
\cmidrule(lr){2-6}\cmidrule(lr){7-11}\cmidrule(lr){12-16}
& \textbf{W1} & \textbf{W2} & \textbf{W3} & \textbf{W4} & \textbf{Avg}
& \textbf{W1} & \textbf{W2} & \textbf{W3} & \textbf{W4} & \textbf{Avg}
& \textbf{W1} & \textbf{W2} & \textbf{W3} & \textbf{W4} & \textbf{Avg} \\
\midrule

\multicolumn{16}{l}{\textbf{Time Series Models}} \\
\midrule

ARIMA~\cite{box2015time}        
& 0.158 & 0.183 & 0.243 & 0.293 & 0.219           
& 0.165 & 0.268 & 0.372 & 0.460 & 0.316          
& 0.152 & 0.183 & 0.261 & 0.320 & 0.229 \\

GRU~\cite{DBLP:journals/corr/ChoMBB14}     
& 0.200 & 0.203 & 0.251 & 0.297 & 0.237 
& 0.208 & 0.266 & 0.316 & 0.346 & 0.284
& 0.229 & 0.186 & 0.252 & 0.301 & 0.243 \\

Autoformer~\cite{wu2021autoformer}
& 0.385 & 0.306 & 0.336 & 0.347 & 0.343
& 0.331 & 0.341 & 0.394 & 0.425 & 0.373
& 0.364 & 0.186 & 0.240 & \textbf{0.277} & 0.267 \\

Crossformer~\cite{zhang2023crossformer}
& 0.214 & 0.215 & 0.249 & 0.287 & 0.241
& 0.145 & 0.203 & 0.274 & 0.326 & 0.237
& 0.169 & 0.174 & 0.244 & 0.300 & 0.222 \\

Timemixer~\cite{wang2023timemixer}
& 0.209 & 0.207 & 0.235 & 0.260 & 0.228
& 0.332 & 0.313 & 0.346 & 0.377 & 0.342
& 0.380 & 0.188 & 0.245 & 0.292 & 0.276 \\
\midrule
\multicolumn{16}{l}{\textbf{Multimodal Time Series Models}} \\
\midrule

Maestro~\cite{maestro2025}
& 0.194 & 0.197 & 0.246 & 0.275 & 0.228
& 0.145 & 0.203 & 0.261 & 0.311 & 0.230
& 0.157 & 0.174 & 0.243 & 0.296 & 0.217 \\

DiPro~\cite{liu2025multimodal}
& 0.200 & 0.192 & 0.241 & 0.279 & 0.228
& 0.155 & 0.217 & 0.265 & 0.315 & 0.238
& 0.178 & 0.181 & 0.247 & 0.297 & 0.226 \\

\midrule

\multicolumn{16}{l}{\textbf{Epidemiological Models}} \\
\midrule
Persistence~\cite{mathis2024flusight}   
& \textbf{0.129} & 0.194 & 0.244 & 0.284 & 0.213
& \textbf{0.138} & 0.217 & 0.278 & 0.328 & 0.240
& \textbf{0.139} & 0.2199 & 0.279 & 0.329 & 0.242\\

CNNRNN-Res~\cite{wu2018deep}  
& 0.247 & 0.286 & 0.330 & 0.370 & 0.308 
& 0.248 & 0.304 & 0.355 & 0.397 & 0.326
& 0.242 & 0.333 & 0.425 & 0.502 & 0.376 \\

Cola-GNN~\cite{deng2020cola}    
& 0.148 & 0.252 & 0.335 & 0.427 & 0.291
& 0.235 &  0.339 & 0.356 & 0.371 & 0.325
& 0.183 & 0.328 & 0.348 & 0.417 & 0.318 \\
\midrule

\multicolumn{16}{l}{\textbf{Multimodal Epidemic Forecasting (Ours)}} \\
\midrule
\textbf{M-SPICE} (TS only)
& 0.192 & 0.189 & 0.231 & 0.260 & 0.218
& 0.144 & 0.201 & 0.254 & 0.310 & 0.228 
& 0.156 & 0.177 & 0.242 & 0.292 & 0.217  \\

\textbf{M-SPICE} (TS+Spatial)
& 0.164 & \textbf{0.167} & \textbf{0.209} & \textbf{0.236} & \textbf{0.194}
& 0.149 & \textbf{0.196} & \textbf{0.249} & \textbf{0.287} & \textbf{0.220}
& 0.142 & \textbf{0.173} & \textbf{0.239} &  0.288 & \textbf{0.210} \\
\bottomrule
\end{tabular}}
\end{table*}

Across all targets, \emph{temporal-only models} exhibit substantial degradation at longer horizons, particularly beyond two weeks (W3–W4). This trend is consistent across classical (ARIMA), neural (GRU), and transformer-based baselines, highlighting the inherent limitations of relying solely on historical surveillance signals for mid-range epidemic forecasting.\\
\indent\emph{Epidemic forecasting models} exhibit mixed performance despite being designed specifically for infectious disease prediction. Notably, the persistence baseline remains highly competitive and outperforms several learned forecasting models across multiple targets and horizons. While Cola-GNN and CNNRNN-Res incorporate region-level spatial structure through graph and spatiotemporal neural architectures, respectively, they do not consistently improve upon persistence or temporal-only baselines, particularly at longer horizons. These results highlight the difficulty of extracting useful spatial information from coarse region-level representations and suggest that stronger spatial inductive biases are needed for robust medium-term epidemic forecasting.

Existing \emph{multimodal time series models} like Maestro and DiPro provide a clear lift over pure temporal baselines by incorporating auxiliary signals.
Incorporating spatially localized environmental and disease burden context through structured multimodal fusion yields consistent improvements across all targets. Our model \textbf{M-SPICE} (TS+Spatial) outperforms its temporal-only counterpart at nearly all horizons, with gains becoming more pronounced at weeks W3 and W4. This indicates that spatial auxiliary information is particularly important for medium-term epidemic forecasting. Moreover, multimodal baselines provide comparable performance at short horizons (W1); however, they fail to sustain accuracy at longer horizons, whereas our approach maintains consistent improvements. Our consistent improvement aligns with the design of our gating mechanism, which allows spatial context to modulate predictions dynamically at each forecast horizon. 

While forecast accuracy is important, epidemic forecasting also requires reliable uncertainty estimates to support public health decision-making. Thus, we \textit{evaluate the probabilistic forecasts} using Cov90 and rWIS. Table~\ref{tab:uncertainty_results} shows that performance varies substantially across COVID, flu, and ILI forecasting tasks, with several baselines achieving strong results on individual tasks but exhibiting inconsistent behavior across tasks. In contrast, our temporal and multimodal variants remain among the strongest performers across tasks demonstrating reliable probabilistic forecasting across diverse surveillance targets. Notably, our models consistently maintain competitive coverage while achieving low interval scores, indicating a favorable balance between calibration, sharpness, and forecast accuracy. These results suggest that the proposed framework generalizes reliably across diseases and forecasting settings, producing probabilistic forecasts that remain stable across prediction targets.

\subsection{RQ2: Robustness Across Settings}
We evaluate the robustness of integrating spatial context by analyzing how performance gains vary across tasks, geographic regions, and forecast horizons. 

\subsubsection{Forecast horizons.}
The benefits of spatial integration increase with forecast horizon as shown in Table~\ref{tab:results_grouped_by_disease}. Improvements are modest at short horizons (W1), where temporal dynamics dominate, but become more pronounced at longer horizons (W3--W4). This pattern suggests that spatial context provides complementary information that is particularly valuable for medium-term forecasting, where temporal signals alone become less predictive.

\subsubsection{Forecast Tasks.}
Table~\ref{tab:results_grouped_by_disease} shows consistent improvements across COVID-19, influenza, and ILI forecasting tasks, indicating that the benefits of spatial integration generalize across disease domains. Despite differences in transmission dynamics and data characteristics, the model is able to leverage spatial structure to improve performance, supporting the robustness of the approach.

\subsubsection{Geographic regions.}
To further examine regional heterogeneity, we analyze the relationship between performance gains and regional characteristics using Pearson correlation and hypothesis testing shown in Table~\ref{tab:region_correlation}. We find that the benefits of integrating spatial context are strongest in regions with greater spatial structure and cross-boundary interaction. In particular, performance gains are higher in states with more bordering neighbors (r=+0.551, p<0.001) and in states with greater sub-regional granularity, as measured by county count (r=+0.430, p=0.002). These results suggest that integrating spatial context is most beneficial in regions where both cross-boundary transmission pathways and fine-grained spatial heterogeneity are present, highlighting that the effectiveness of spatial context is not uniform but instead depends on the underlying geographic and epidemiological structure of the region. 

\subsubsection{Robustness to spatial input structure.}
To further validate the role of sub-regional spatial context, we conduct an ablation analysis on the spatial inputs. Specifically, we evaluate three settings: dynamic COVID-19 maps, static maps fixed to a representative week, and no spatial input, using the trained influenza model. Results show that dynamic maps yield the best performance, followed by static maps, while removing spatial input leads to the largest degradation across all horizons. This ordering indicates that both the presence of spatial context and its time-varying structure contribute meaningfully to forecasting performance, and that the learned spatial representations generalize across disease domains. Detailed ablation results are reported in Appendix~\ref{app:spatial_ablation}.

Overall, these findings demonstrate that integrating spatial context yields robust improvements across settings, while also exhibiting meaningful variation that reflects differences in temporal dynamics, disease characteristics, and regional structure.
\begin{table}[ht]
\centering
\captionsetup{
    skip=4pt
}
\caption{Evaluation of probabilistic forecasts for 1--4 week ahead prediction of COVID and flu hospitalizations and ILI percentage. Lower rWIS and Cov90 values closer to 0.90 indicate better uncertainty estimates. Second best is underlined.}

\label{tab:uncertainty_results}

\resizebox{\columnwidth}{!}{
\vspace{-10pt}
\setlength{\tabcolsep}{4pt}
\renewcommand{\arraystretch}{}
\begin{tabular}{l|cc|cc|cc}
\toprule
\multirow{2}{*}{\textbf{Model}} &
\multicolumn{2}{c|}{\textbf{COVID hosp}} &
\multicolumn{2}{c|}{\textbf{flu hosp}} &
\multicolumn{2}{c}{\textbf{ILI percentage}} \\
\cmidrule(lr){2-3}
\cmidrule(lr){4-5}
\cmidrule(lr){6-7}
& \textbf{Cov90} & \textbf{rWIS}
& \textbf{Cov90} & \textbf{rWIS}
& \textbf{Cov90} & \textbf{rWIS} \\
\midrule


ARIMA~\cite{box2015time}
& 0.781 & 1.144
& 0.668 & 1.465
& 0.734 & 1.261 \\

GRU~\cite{DBLP:journals/corr/ChoMBB14}
& \textbf{0.846} & 0.963
& \underline{0.771} & 1.026
& \textbf{0.794} & 1.194 \\

Autoformer~\cite{wu2021autoformer}
& \underline{0.816} & 1.694
& 0.716 & 1.810
& 0.747 & 1.716 \\

Crossformer~\cite{zhang2023crossformer}
& 0.806 & 1.221
& 0.699 & 1.035
& 0.742 & 1.004 \\

Timemixer~\cite{wang2023timemixer}
& 0.813 & 0.993
& 0.706 & 1.564
& 0.728 & 1.748 \\

\midrule

Maestro~\cite{maestro2025}
& 0.792 & 1.067
& 0.734 & 0.799
& \underline{0.787} & \underline{0.934} \\

DiPro~\cite{liu2025multimodal}
& 0.772 & 1.102
& 0.753 & 0.834
& 0.773 & 1.025 \\

\midrule
Persistence~\cite{mathis2024flusight}
& 0.798 & 1.000
& 0.743 & 1.000
& 0.760 & 1.000 \\

CNNRNN-Res~\cite{wu2018deep}
& 0.804 & \underline{0.686}
& 0.675 & 0.775
& 0.743 & 1.428 \\

Cola-GNN~\cite{deng2020cola}
& 0.812 & \textbf{0.647}
& 0.672 & \textbf{0.663}
& 0.705 & 1.242 \\

\midrule

\textbf{M-SPICE} (TS only)
& 0.787 & 0.921
& \textbf{0.776} & \underline{0.745}
& 0.784 & 0.947 \\

\textbf{M-SPICE} (TS+Spatial)
& 0.801 & 0.856
& 0.755 & 0.801
& 0.783 & \textbf{0.900} \\

\bottomrule
\end{tabular}
}
\end{table}

\subsection{RQ3: Interpretability and Scientific Insight}
To investigate what spatial context is leveraged by the model and what it reveals about disease dynamics, we analyze attention patterns from the joint self-attention fusion module. 

\begin{table}[htbp]
\centering
\captionsetup{
    skip=4pt
}
\caption{
Attention mass aggregated by modality direction and normalized relative to uniform attention,
with regions grouped by forecasting performance (Best-, Mid-, and Worst-3) ranked by mean NRMSE.
Values greater than $1\times$ indicate amplification relative to uniform routing.
}
\resizebox{0.9\linewidth}{!}{
\small
\setlength{\tabcolsep}{6pt}
\begin{tabular}{llccccc}
\toprule
Group & Region 
& ts$\rightarrow$ts 
& ts$\rightarrow$img 
& img$\rightarrow$ts 
& img$\rightarrow$img 
& ts$\leftrightarrow$img  \\
\midrule
Best-3 
& IL & 1.22$\times$ & 0.78$\times$ & 1.85$\times$ & 0.15$\times$ & 2.63$\times$ \\
& IA & 1.28$\times$ & 0.72$\times$ & 1.82$\times$ & 0.18$\times$ & 2.54$\times$ \\
& KY & 1.24$\times$ & 0.76$\times$ & 1.68$\times$ & 0.32$\times$ & 2.43$\times$ \\
\midrule
Mid-3 
& WI & 1.17$\times$ & 0.83$\times$ & 1.83$\times$ & 0.17$\times$ & 2.66$\times$ \\
& CA & 1.27$\times$ & 0.73$\times$ & 1.96$\times$ & 0.04$\times$ & 2.69$\times$ \\
& PA & 1.19$\times$ & 0.81$\times$ & 1.83$\times$ & 0.17$\times$ & 2.64$\times$ \\
\midrule
Worst-3 
& DE & 1.11$\times$ & 0.89$\times$ & 1.97$\times$ & 0.03$\times$ & 2.86$\times$ \\
& GA & 1.08$\times$ & 0.92$\times$ & 1.70$\times$ & 0.30$\times$ & 2.62$\times$ \\
& HI & 1.24$\times$ & 0.76$\times$ & 1.98$\times$ & 0.02$\times$ & 2.74$\times$ \\
\bottomrule
\end{tabular}
}
\vspace{-5pt}
\label{tab:attn_x_uniform}
\end{table}
\vspace{-5pt}

\subsubsection{Performance-stratified attention patterns.}
Table~\ref{tab:attn_x_uniform} summarizes attention mass aggregated by modality direction and normalized relative to uniform routing, with regions grouped by forecasting performance. Across all performance tiers, image-to-time-series attention (\texttt{img$\rightarrow$img}) is consistently amplified ($1.65\times$--$1.98\times$), while time-series-to-image (\texttt{ts$\rightarrow$ts}) and image-to-image (\texttt{img$\rightarrow$img}) attention are suppressed ($<1\times$). This pattern indicates that spatial context is primarily used to condition and refine temporal representations rather than drive predictions independently. Finally, cross-modal interactions (\texttt{img$\leftrightarrow$ts}) are stronger than within-modality interactions, indicating that auxiliary spatial inputs actively support and modulate epidemic time-series representations. Motivated by these global patterns, we next examine California as a representative mid-performing region to analyze how cross-modal attention evolves over the epidemic season.

\subsubsection{Directional multimodal attention dynamics.}
\begin{figure}[htbp]
\centering
\includegraphics[width=0.9\linewidth]{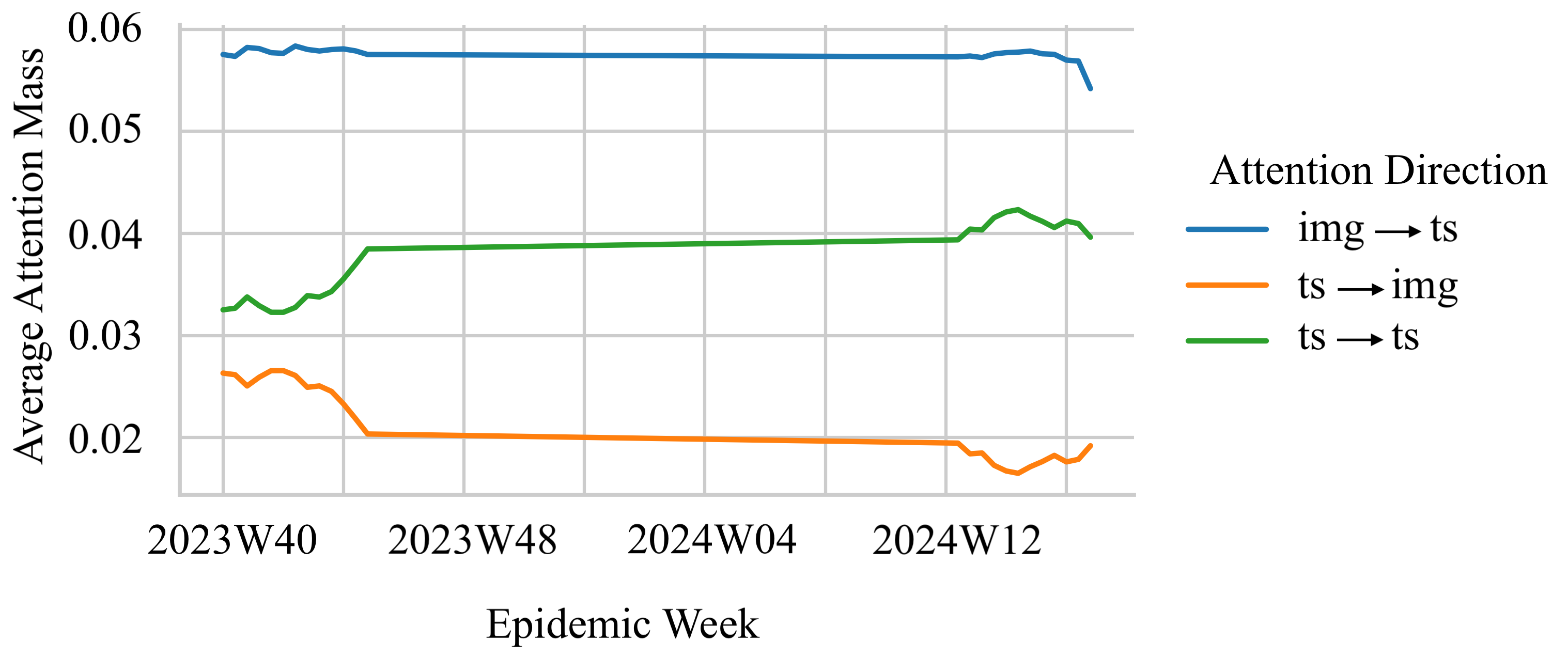}
\caption{
Directional multimodal attention dynamics in California. Image tokens consistently attend to temporal tokens (img$\rightarrow$ts), while temporal-to-image attention (ts$\rightarrow$img) remains low, suggesting that spatial information is primarily used to condition and refine temporal representations rather than drive predictions independently.
}
\vspace{-10pt}
\label{fig:cross_model_attn_mass}
\end{figure}

Figure~\ref{fig:cross_model_attn_mass} shows the evolution of cross-modal attention mass over time for California. We observe a consistent asymmetry in cross-modal interactions: image tokens attend strongly to temporal tokens (img→ts), while temporal tokens exhibit comparatively weak attention to image tokens (ts→img). At the same time, temporal self-attention (ts→ts) remains substantial throughout the season. This pattern suggests that spatial context is primarily used to condition and interpret temporal disease dynamics, rather than replacing or overwhelming them. Such directional fusion aligns with epidemiological reasoning, where spatial signals constrain short- and medium-term temporal trends without dominating temporal inference. While Figure~\ref{fig:cross_model_attn_mass} summarizes directional attention dynamics aggregated over time, it does not reveal how spatial context is selected or aligned at specific points in the epidemic season. To examine how these asymmetric interactions manifest at the token level, we visualize joint attention patterns and corresponding spatial inputs for California across early, mid, and late seasonal stages.

\begin{figure}[htbp]
    \centering
    \includegraphics[width=0.9\linewidth]{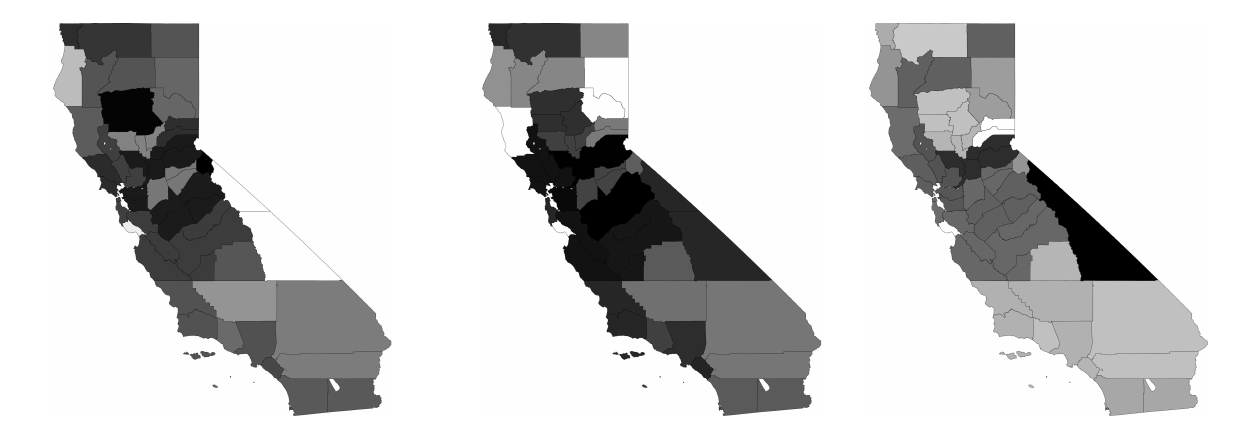}
    \vspace{0.6em}
    \includegraphics[width=\linewidth]{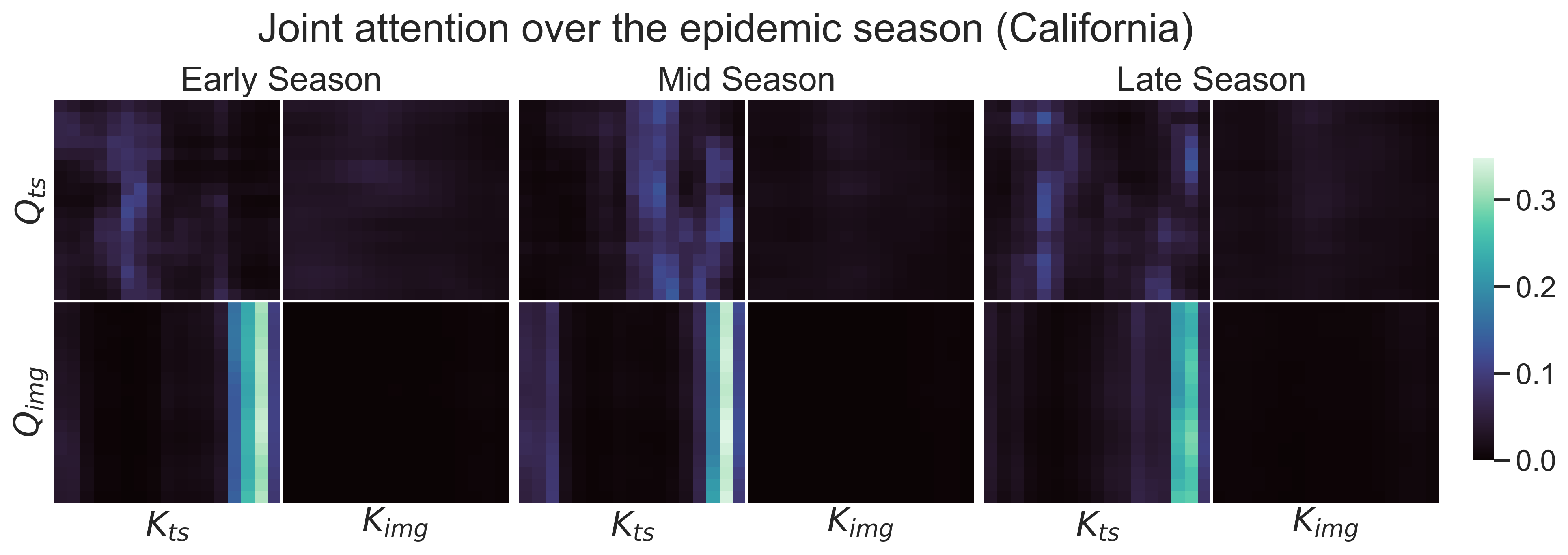}
    \caption{Disease burden proxy and joint attention patterns for California across the epidemic season.
    From left to right, panels correspond to early (2023W40), mid (2024W06), and late (2024W12) stages of the epidemic season. For each stage, there is a county-level disease burden proxy map (top) and the corresponding joint self-attention heat maps (bottom).}
    \label{fig:attn_ca_seasonal}
\end{figure}
\vspace{-3pt}
\subsubsection{Seasonal variation in spatial context utilization.}
Figure~\ref{fig:attn_ca_seasonal} illustrates the evolution of joint self-attention patterns for California across the epidemic season, shown at early (2023W40), mid (2024W06), and late (2024W12) stages. Attention maps are averaged over heads and computed over interleaved temporal and image tokens to inspect cross-modal interactions. Spatial attention is present even in the early season, but reaches its highest intensity during the mid-season period, when disease burden becomes most spatially structured, before becoming more localized in the late season. This pattern indicates that spatial context is leveraged throughout the season, with its influence becoming most pronounced during periods of active and spatially organized transmission, when regional signals provide meaningful constraints on temporal forecasting.

This finding aligns with prior work showing that temporal signals dominate epidemic onset and short-term dynamics, while environmental and spatial factors act as slower-moving constraints on epidemic evolution \cite{ruan2025climate, choo2026measuring}.  By transitioning from temporally dominant attention to structure-aware multimodal fusion, our framework captures this later stage of epidemic evolution, where reliance on atomic time-series signals alone becomes insufficient.

\subsection{Michigan Case Study: Disease-Matched vs. Disease Burden Proxy Maps}
In the primary experiments, county-level COVID-19 hospitalization maps are used as a shared proxy for spatial disease burden across forecasting targets. To evaluate the potential benefit of using disease-matched and temporally aligned spatial information instead, we conduct a Michigan case study using county-level maps constructed from data provided by the Michigan Department of Health and Human Services (MDHHS). Specifically, COVID-19 forecasts use COVID-19 hospitalization maps, whereas influenza forecasts use influenza hospitalization maps. Table~\ref{tab:mi_rich_vs_old} shows that for COVID hospitalization forecasting, disease-matched maps consistently improve performance across all forecast horizons, reducing the average NRMSE from $0.191$ to $0.176$. In contrast, performance for influenza hospitalization forecasting is comparable between disease-matched and proxy maps, with proxy maps slightly outperforming disease-matched maps on average ($0.261$ vs.\ $0.263$). These results suggest that proxy spatial signals capture much of the spatial information relevant for forecasting and provide a practical alternative when disease-matched, temporally aligned spatial surveillance data are unavailable. However, disease-matched and temporally aligned maps can yield substantial improvements, as observed for COVID hospitalization forecasting, highlighting the potential value of richer spatial information when available.
\begin{table}[htbp]
\centering
\small
\setlength{\tabcolsep}{4pt}
\caption{Average NRMSE over seeds $= [5,17,33]$ for Michigan county-level maps comparing disease-matched v. proxy}
\label{tab:mi_rich_vs_old}
\resizebox{\columnwidth}{!}{
\begin{tabular}{l|ccccc|ccccc}
\toprule
& \multicolumn{5}{c|}{\textbf{COVID hospitalizations}}
& \multicolumn{5}{c}{\textbf{influenza hospitalizations}} \\
\cline{2-11}
\textbf{County-level Maps}
& \textbf{W1} & \textbf{W2} & \textbf{W3} & \textbf{W4} & \textbf{Avg}
& \textbf{W1} & \textbf{W2} & \textbf{W3} & \textbf{W4} & \textbf{Avg} \\
\midrule

Disease-Matched
& \textbf{0.155} & \textbf{0.157} & \textbf{0.185} & \textbf{0.207} & \textbf{0.176}
& \textbf{0.164} & 0.238 & \textbf{0.314} & 0.334 & 0.263 \\

Disease Burden Proxy
& 0.162 & 0.170 & 0.204 & 0.229 & 0.191
& 0.165 & \textbf{0.236} & 0.315 & \textbf{0.327} & \textbf{0.261} \\

\bottomrule
\end{tabular}
}
\end{table}